\documentclass{llncs}
\usepackage{llncsdoc}
\usepackage{times}
\usepackage{graphicx}
\usepackage{float}
\usepackage{amsmath}
\usepackage{amsfonts}
\usepackage{rotating}
\usepackage[ruled,vlined]{algorithm2e}
\usepackage{algorithmic}
\usepackage{multirow}
\usepackage{multicol}
\usepackage{blindtext}

\makeatletter

\newcommand{\Rmnum}[1]{\expandafter\@slowromancap\romannumeral #1@}
\makeatother

\begin{document}

\title{Multi-label Learning via Structured Decomposition and Group Sparsity}
\author{Tianyi Zhou \and Dacheng Tao}
\institute{Centre for Quantum Computation and Intelligent Systems, Faculty of Engineering and Information Technology, University of Technology Sydney, Broadway NSW 2007, Australia. \email{dacheng.tao@uts.edu.au}}
\maketitle

\begin{abstract}
In multi-label learning, each sample is associated with several labels. Existing works indicate that exploring correlations between labels improve the prediction performance. However, embedding the label correlations into the training process significantly increases the problem size. Moreover, the mapping of the label structure in the feature space is not clear. In this paper, we propose a novel multi-label learning method ``Structured Decomposition + Group Sparsity (SDGS)''. In SDGS, we learn a feature subspace for each label from the structured decomposition of the training data, and predict the labels of a new sample from its group sparse representation on the multi-subspace obtained from the structured decomposition. In particular, in the training stage, we decompose the data matrix $X\in R^{n\times p}$ as $X=\sum_{i=1}^kL^i+S$, wherein the rows of $L^i$ associated with samples that belong to label $i$ are nonzero and consist a low-rank matrix, while the other rows are all-zeros, the residual $S$ is a sparse matrix. The row space of $L_i$ is the feature subspace corresponding to label $i$. This decomposition can be efficiently obtained via randomized optimization. In the prediction stage, we estimate the group sparse representation of a new sample on the multi-subspace via group \emph{lasso}. The nonzero representation coefficients tend to concentrate on the subspaces of labels that the sample belongs to, and thus an effective prediction can be obtained. We evaluate SDGS on several real datasets and compare it with popular methods. Results verify the effectiveness and efficiency of SDGS.
\end{abstract}

\section{Introduction}

Multi-label learning \cite{LearningFML} aims to find a mapping from the feature space $\mathcal X\subseteq \mathbb R^p$ to the label vector space $\mathcal Y\subseteq \{0,1\}^k$, wherein $k$ is the number of labels and $y_i=1$ denotes the sample belongs to label $i$. Binary relevance (BR) \cite{MLreview2} and label powerset (LP) \cite{MLreview2} are two early and natural solutions. BR and LP transform a multi-label learning problem to several binary classification tasks and single-label classification task, respectively. Specifically, BR associates each label with an individual class, i.e., assigns samples with the same label to the same class. LP treats each unique set of labels as a class, in which samples share the same label vector.

Although BP/LP and their variants can directly transform a multi-label learning problem into multiple binary classification tasks or single-label classification task, multi-label learning brings new problems. First, the labels are not mutually exclusive in multi-label learning, and thus it is necessary to consider not only the discriminative information between different labels but also their correlations. Second, the large number of labels always leads to the imbalance between positive samples and negative ones in each class, and this limits the performance of binary classification algorithms. Third, the problem size of multi-label learning will be significantly increased when it is decomposed into many binary classification problems.

Recent multi-label learning methods more or less tackle some of the above problems and demonstrate that the prediction performance can be improved by exploiting specific properties of the multi-label data, e.g., label dependence, label structure, and the dependence between samples and the corresponding labels. We categorize popular methods into two groups.

\begin{enumerate}
\item The first group of methods transform multi-label prediction into a sequence of binary classification methods with special structures implied by label correlations. For example, the random k-labelsets (RAkEL) method \cite{RAkEL} randomly selects an ensemble of subset from the original labelsets, and then LP is applied to each subset. The final prediction is obtained by ranking and thresholding of the results on the subsets. Hierarchical binary relevance (HBR) \cite{HBR} builds a general-to-specific tree structure of labels, where a sample with a label must be associated with its parent labels. A binary classifier is trained on each non-root label. Hierarchy of multi-label classifiers (HOMER) \cite{HOMER} recursively partitions the labels into several subsets and build a tree-shaped hierarchy. A binary classifier is trained on each non-root label subset. The classifier chain (CC) \cite{CC} adopts a greedy way to predict unknown label from feature and predicted labels via binary classifier.
\item The second group of methods formulate the multi-label prediction to other kinds of problems rather than binary classification. For example, the C\&W procedure \cite{CandW} separates the problem into two stages, i.e., BR and correction of the BR results by using label dependence. Regularized multi-task learning \cite{YeLowrankSparse} and shared-subspace learning \cite{SharedSubspace} formulate the problem as regularized regression or classification problem. Multi-label k-nearest neighbor (ML-kNN) \cite{MLKNN} is an extension of kNN. Multi-label dimensionality reduction via dependence maximization (MDDM) \cite{MDDM} maximizes the dependence between feature space and label space, and provides a data preprocessing for other multi-label learning method. A linear dimensionality reduction method for multi-label data is proposed in \cite{JiML}. In \cite{MultiLabelCS}, multi-label prediction is formulated as a sparse signal recovery problem.
\end{enumerate}

However, the problem size always significantly increases when multi-label learning are decomposed into a set of binary classification problems or formulated as another existing problem, because the label correlations need to be additionally considered. Furthermore, the mapping of label structure in feature space has not been studied. In this paper, we propose a novel multi-label learning method ``Structured Decomposition + Group Sparsity (SDGS)'', which assigns each label a corresponding feature subspace via randomized decomposition of the training data, and predicts the labels of a new sample by estimating its group sparse representation in the obtained multi-subspace.

In the training stage, SDGS approximately decomposes the data matrix $X\in\mathbb R^{n\times p}$ (each row is a training sample) as $X=\sum_{i=1}^kL^i+S$. In the matrix $L^i$, only the rows corresponding to samples with label $i$ (i.e., $y_i=1$) are nonzero. These rows represent the components determined by label $i$ in the samples and compose a low-rank matrix, which row space is the feature subspace characterized by label $i$. The matrix $S$ represents the residual components that cannot be explained by the given labels and is constrained to be sparse. The decomposition is obtained via a randomized optimization with low time complexity.

In the prediction stage, SDGS estimates the group sparse representation of a new sample in the obtained multi-subspace via group \emph{lasso} \cite{GroupLasso}. The representation coefficients associated with basis in the same subspace are in the same group. Since the components caused by a specific label can be linearly represented by the corresponding subspace obtained in the training stage, the nonzero representation coefficients will concentrate on the groups corresponding to the labels that the sample belongs to. This gives the rational of the proposed SDGS for multi-label learning. Group \emph{lasso} is able to select these nonzero coefficients group-wisely and thus the labels can be identified.

SDGS provides a novel and natural multi-label learning method by building a mapping of label structure in decomposed feature subspaces. Group sparse representation in the multi-subspace is applied to recover the unknown labels. It embeds the label correlations without increasing the problem size and is robust to the imbalance problem. By comparing SDGS with different multi-label learning methods, we show its effectiveness and efficiency on several datasets.

\section{Assumption and Motivation}

Given a sample $x\in\mathbb R^p$ and its label vector $y\in\{0,1\}^k$, we assume that $x$ can be decomposed as the sum of several components $l^i$ and a sparse residual $s$:
\begin{equation}\label{E:model}
x=\sum\limits_{i:y_i=1}l^i+s.
\end{equation}
The component $l^i$ is caused by the label $i$ that $x$ belongs to. Thus $l^i$ can be explained as the mapping of label $i$ in $x$. The residual $s$ is the component that all the labels in $y$ cannot explain. The model in (\ref{E:model}) reveals the general relationship between feature space and labels.

For all the samples with label $i$, we assume their components corresponding to label $i$ lies in a linear subspace $C^i\in\mathbb R^{r^i\times p}$, i.e., $l^i=\beta_{G_i}C^i$, wherein $\beta_{G_i}$ is the representation coefficients corresponding to $C^i$. Thus the model (\ref{E:model}) can be equivalently written as:
\begin{equation}\label{E:modell}
\begin{array}{ll}
x&=\sum\limits_{i=1}^k\beta_{G_i}C^i+s,\\
&\forall i\in\{i:y_i=0\},\beta_{G_i}=\textbf{0}.
\end{array}
\end{equation}
If we build a dictionary $C=[C^1;\dots;C^k]$ as the multi-subspace characterized by the $k$ labels, the corresponding representation coefficient vector for $x$ is $\beta=[\beta_{G_1},\dots,\beta_{G_k}]$. The coefficients $\beta_{G_i}$ corresponding to the labels $x$ does not belong to are zeros, so $\beta$ is group sparse, wherein the groups are $G_i,i=1,\dots,k$.

In training stage of SDGS, we learn the multi-subspace $C^i,i=1,\dots,k$ from the training data via a structured decomposition, in which the components corresponding to label $i$ from all the samples consists a low-rank matrix $L^i_{\Omega_i}$, wherein $\Omega_i$ is the index set of samples with label $i$. Thus the row space of $L^i_{\Omega_i}$ is the subspace $C^i$. In the prediction stage of SDGS, given a new sample $x$, we apply group \emph{lasso} to find the group sparse representation $\beta$ on the multi-subspace $C$, and then a simple thresholding is used to test which groups $\beta$ concentrates on. The labels that these groups corresponds to are predicted labels for the sample $x$.

In the training stage, the label correlations and structure are naturally preserved in their mappings $C^i$. In the prediction stage, both discriminative and structured information encoded in labels are considered via group \emph{lasso}. Therefore, SDGS explores label correlations without increasing the problem size.

\vspace{-2mm}
\section{Training: Structured Decomposition}

In this section, we introduce the training stage of SDGS, which approximately decomposes the data matrix $X\in\mathbb R^{n\times p}$ into $X=\sum_{i=1}^kL^i+S$. For the matrix $L^i$, the rows corresponding to the samples with label $i$ are nonzero, while the other rows are all-zero vectors. The nonzero rows are the components caused by label $i$ in the samples. We use $\Omega_i$ to denote the index set of samples with label $i$ in the matrix $X$ and $L^i$, and then the matrix composed of the nonzero rows in $L^i$ is represented by $L^i_{\Omega_i}$. In the decomposition, the rank of $L^i_{\Omega_i}$ is upper bounded, which indicates that all the components caused by label $i$ nearly lies in a linear subspace. The matrix $S$ is the residual of the samples that cannot be explained by the given labels. In the decomposition, the cardinality of $S$ is upper bounded, which makes $S$ sparse.

If the label matrix of $X$ is $Y\in\{0,1\}^{n\times k}$, the rank of $L^i_{\Omega_i}$ is bounded not more than $r^i$ and the cardinality of $S$ is bounded not more than $K$, the decomposition can be written as solving the following constrained minimization problem:
\begin{equation}\label{E:ms}
\begin{array}{rl}
\min\limits_{L^i,S}&\left\|X-\sum_{i=1}^kL^i-S\right\|_F^2\\
s.t.&{\rm rank}\left(L^i_{\Omega_i}\right)\leq r^i,L^i_{\overline{\Omega}_i}=\textbf{0},\forall i=1,\dots,k\\
&{\rm card}\left(S\right)\leq K.
\end{array}
\end{equation}
Therefore, each training sample in $X$ is decomposed as the sum of several components, which respectively correspond to several labels that the sample belongs to. SDGS separates these components from the original sample by building the mapping of $Y$ in the feature space of $X$. For label $i$, we obtain its mapping in the feature subspace as the row space of $L^i_{\Omega_i}$.

\vspace{-2mm}
\subsection{Alternating minimization}

Although the rank constraint to $L^i_{\Omega_i}$ and cardinality constraint to $S$ are not convex, the optimization in (\ref{E:ms}) can be solved by alternating minimization that decomposes it as the following $k+1$ subproblems, each of which has the global solutions:
\begin{equation}\label{E:mssub}
\left\{
  \begin{array}{ll}
    L^i_{\Omega_i}=\arg\min\limits_{{\rm rank}\left(L^i_{\Omega_i}\right)\leq r^i}\left\|X-\sum\limits_{j=1,j\neq i}^kL^j-S-L^i\right\|_F^2, \\
    ~~~~~~~~\forall i=1,\dots,k.\\
    S=\arg\min\limits_{{\rm card}\left(S\right)\leq K}\left\|X-\sum\limits_{j=1}^kL^j-S\right\|_F^2.
  \end{array}
\right.
\end{equation}

The solutions of $L^i_{\Omega_i}$ and $S$ in above subproblems can be obtained via hard thresholding of singular values and the entries, respectively. Note that both SVD and matrix hard thresholding have global solutions. In particular, $L^i_{\Omega_i}$ is built from the first $r^i$ largest singular values and the corresponding singular vectors of $\left(X-\sum_{j=1,j\neq i}^kL^j-S\right)_{\Omega_i}$, while $S$ is built from the $K$ entries with the largest absolute value in $X-\sum_{j=1}^kL^j$, i.e,
\begin{equation}\label{E:mssolution}
\left\{
  \begin{array}{ll}
    L^i_{\Omega_i}=\sum\limits_{q=1}^{r^i}\lambda_qU_qV_q^T, i=1,\dots,k,\\
    {\rm svd}\left[\left(X-\sum_{j=1,j\neq i}^kL^j-S\right)_{\Omega_i}\right]=U\Lambda V^T; \\
    S=\mathcal {P}_{\Phi}\left(X-\sum\limits_{j=1}^kL^j\right), \Phi:\left|\left(X-\sum\limits_{j=1}^kL^j\right)_{{r,s}\in{\Phi}}\right|\neq0 \\ {\rm~and~} \geq \left|\left(X-\sum\limits_{j=1}^kL^j\right)_{{r,s}\in{\overline{\Phi}}}\right|, |\Phi|\leq K.
  \end{array}
\right.
\end{equation}
The projection $S=\mathcal {P}_{\Phi}(R)$ represents that the matrix $S$ has the same entries as $R$ on the index set $\Phi$, while the other entries are all zeros.

The decomposition is then obtained by iteratively solving these $k+1$ subproblems in (\ref{E:mssub}) according to (\ref{E:mssolution}). In this paper, we initialize $L^i_{\Omega_i}$ and $S$ as
\begin{equation}\label{E:msinitial}
\left\{
  \begin{array}{ll}
  L^i_{\Omega_i}:=Z_{\Omega_i},i=1,\dots,k,\\
  Z=D^{-1}X,D={\rm diag}\left(Y\textbf{1}\right);\\
  S:=\textbf{0}.
  \end{array}
\right.
\end{equation}
In each subproblem, only one variable is optimized with the other variables fixed. The convergence of this alternating minimization can be proved in Theorem \ref{T:ls_convergence} by demonstrating that the approximation error keeps monotonically decreasing throughout the algorithm.

\begin{theorem}\label{T:ls_convergence}
The alternating minimization of subproblems (\ref{E:mssub}) produces a sequence of $\|X-\sum_{i=1}^kL^i-S\|_F^2$ that converges to a local minimum.
\end{theorem}
\begin{proof}
Let the objective value (decomposition error) $\|X-\sum_{i=1}^kL^i-S\|_F^2$ after solving the $k+1$ subproblems in (\ref{E:mssub}) to be $E^1_{(t)},\dots,E^{k+1}_{(t)}$ respectively for the $t^{th}$ iteration round. We use subscript $(t)$ to signify the variable that is updated in the $t^{th}$ iteration round. Then $E^1_{(t)},\dots,E^{k+1}_{(t)}$ are
\begin{align}
&E^1_{(t)}=\left\|X-S_{(t-1)}-L^1_{(t)}-\sum_{i=3}^kL^i_{(t-1)}-L^2_{(t-1)}\right\|_F^2,\\
&E^2_{(t)}=\left\|X-S_{(t-1)}-L^1_{(t)}-\sum_{i=3}^kL^i_{(t-1)}-L^2_{(t)}\right\|_F^2,\\
\notag&~~~~~~~~~~~~~~~~~~~~~~~~\vdots\\
&E^k_{(t)}=\left\|X-\sum_{i=1}^kL^i_{(t)}-S_{(t-1)}\right\|_F^2,\\
&E^{k+1}_{(t)}=\left\|X-\sum_{i=1}^kL^i_{(t)}-S_{(t)}\right\|_F^2,
\end{align}
The global optimality of $L^i_{(t)}$ yields $E^1_{(t)}\geq E^2_{(t)}\geq\cdots\geq E^k_{(t)}$. The global optimality of $S_{(t)}$ yields $E^k_{(t)}\geq E^{k+1}_{(t)}$. In addition, we have
\begin{align}
&E^{k+1}_{(t)}=\left\|X-\sum_{i=2}^kL^i_{(t)}-S_{(t)}-L^1_{(t)}\right\|_F^2, \\
&E^1_{(t+1)}=\left\|X-\sum_{i=2}^kL^i_{(t)}-S_{(t)}-L^1_{(t+1)}\right\|_F^2.
\end{align}
The global optimality of $L^1_{(t+1)}$ yields $E^{k+1}_{(t)}\geq E^1_{(t+1)}$. Therefore, the objective value (or the decomposition error) $\|X-\sum_{i=1}^kL^i-S\|_F^2$ keeps decreasing throughout the iteration rounds of (\ref{E:mssolution}), i.e.,
\begin{equation}\label{E:converge}
E^1_{(1)}\geq E^{k+1}_{(1)}\geq \cdots\geq E^1_{(t)}\geq E^{k+1}_{(t)}\geq\cdots
\end{equation}
Since the objective value of (\ref{E:ms}) is monotonically decreasing and the constraints are satisfied all the time, iteratively solving (\ref{E:mssub}) produces a sequence of objective values that converge to a local minimum. This completes the proof.
\end{proof}

After obtaining the decomposition by solving (\ref{E:ms}), each training sample is represented by the sum of several components in $L^i$ characterized by the labels it belongs to and the residual in $S$. Therefore, the mapping of label $i$ in feature subspace is defined as the row space $C^i\in\mathbb R^{{r^i}\times p}$ of the matrix $L^i_{\Omega_i}$, which can be obtained via the QR decomposition of $\left(L^i_{\Omega_i}\right)^T$.

\subsection{Accelerate SDGS via bilateral random projections}

The main computation in (\ref{E:mssolution}) is the $k$ times of SVD in obtaining $L^i_{\Omega_i}(i=1,\dots,k)$. SVD requires $\min\left(mn^2,m^2n\right)$ flops for an $m\times n$ matrix, and thus it is impractical when $X$ is of large size. Random projection is effective in accelerating the matrix multiplication and decomposition \cite{RandomSVD}. In this paper, we introduce ``bilateral random projections (BRP)'', which is a direct extension of random projection, to accelerate the optimization of $L^i_{\Omega_i}(i=1,\dots,k)$.

For clear representation, we use letters independent to the ones we use in other parts of this paper to illustrate BRP. In particular, given $r$ bilateral random projections (BRP) of an $m\times n$ dense matrix $X$ (w.l.o.g, $m\geq n$), i.e., $Y_1=XA_1$ and $Y_2=X^TA_2$, wherein $A_1\in\mathbb R^{n\times r}$ and $A_2\in\mathbb R^{m\times r}$ are random matrices,
\begin{equation}\label{E:lr_app}
L=Y_1\left(A_2^TY_1\right)^{-1}Y_2^T
\end{equation}
is a fast rank-$r$ approximation of $X$. The computation of $L$ includes an inverse of an $r\times r$ matrix and three matrix multiplications. Thus, for a dense $X$, $2mnr$ floating-point operations (flops) are required to obtain BRP, $r^2(2n+r)+mnr$ flops are required to obtain $L$. The computational cost is much less than the SVD based approximation.

We build the random matrices $A_1$ and $A_2$ in an adaptive way. Initially, both $A_1$ and $A_2$ are set to standard Gaussian matrices whose entries are independent variables following standard normal distributions. We firstly compute $Y_1=XA_1$, update $A_2:=Y_1$ and calculate the left random projection as $Y_2=X^TA_2$ by using the new $A_2$, and then we update $A_1:=Y_2$ and calculate the right random projection $Y_1=XA_1$ by using the new $A_1$. This adaptive updating of random matrices requires additional flops of $mnr$.

Algorithm 1 summarizes the training stage of SDGS with BRP based acceleration.

\begin{algorithm}[htb]
\SetAlgoLined
\KwIn{$X$, $\Omega_i$, $r^i,i=1,\dots,k$, $K$, $\epsilon$}
\KwOut{$C^i,i=1,\dots,k$}
Initialize $L^i$ and $S$ according to (\ref{E:msinitial}), $t:=0$\;
\While{$\left\|X-\sum_{j=1}^kL^j-S\right\|_F^2>\epsilon$}{
$t:=t+1$\;
\For{$i\leftarrow 1$ \KwTo $k$}{
$N:=\left(X-\sum_{j=1,j\neq i}^kL^j-S\right)_{\Omega_i}$\;
Generate standard Gaussian matrix $A_1\in\mathbb R^{p\times{r^i}}$\;
$Y_1:=NA_1$, $A_2:=Y_1$\;
$Y_2:=N^TY_1$, $Y_1:=NY_2$\;
$L^i_{\Omega_i}:=Y_1\left(A_2^TY_1\right)^{-1}Y_2^T, L^i_{\overline{\Omega}_i}:=\textbf{0}$\;
}
$N:=\left|X-\sum_{j=1}^kL^j\right|$\;
$S:=\mathcal {P}_{\Phi}\left(N\right)$, $\Phi$ is the index set of the first $K$ largest entries of $\left|N\right|$\;
}
QR decomposition $\left(L^i_{\Omega_i}\right)^T=Q^iR^i$ for $i=1,\dots,k$, $C^i:=\left(Q^i\right)^T$\;
\caption{SDGS Training}
\end{algorithm}\vspace{-2mm}

\section{Prediction: Group Sparsity}

In this section, we introduce the prediction stage of SDGS by estimating a group sparse representations of a given sample. In the training stage, we decompose the training data into the sum of low-rank components $L^i_{\Omega_i}$ characterized by their labels and a sparse residual $S$. The mapping of label $i$ in the feature subspace is defined as the row space $C^i$ of $L^i_{\Omega_i}$, because the components of the training data characterized by label $i$ lies in the linear subspace $C^i$.

In the prediction stage of SDGS, we use group \emph{lasso} \cite{GroupLasso} to estimate the group sparse representation $\beta\in\mathbb R^{\sum {r^i}}$ of a test sample $x\in\mathbb R^p$ on the multi-subspace $C=[C^1;\dots;C^k]$, wherein the $k$ groups are defined as index sets of the coefficients corresponding to $C^1,\dots,C^k$. Since group \emph{lasso} selects nonzero coefficients group-wisely, nonzero coefficients in the group sparse representation will concentrate on the groups corresponding to the labels that the sample belongs to.

According to above analysis, we solve the following group \emph{lasso} problem in the prediction of SDGS:
\begin{equation}\label{E:mspredict}
\min\limits_\beta \frac{1}{2}\left\|x-\beta C\right\|_F^2+\lambda\sum\limits_{i=1}^k\left\|\beta_{G_i}\right\|_2,\\
\end{equation}
where the index set $G_i$ includes all the integers between $1+\sum_{j=1}^{i-1}r^j$ and $\sum_{j=1}^{i}r^j$ (including these two numbers).

To obtain the final prediction of label vector $y\in\{0,1\}^k$ for the test sample $x$, we use a simple thresholding of the magnitude sum of coefficients in each group to test which groups that the sparse coefficients in $\beta$ concentrate on:
\begin{equation}\label{E:thresh}
y_\Psi=\textbf{1},y_{\overline\Psi}=\textbf{0},\Psi=\left\{i:\left\|\beta_{G_i}\right\|_1\geq\delta\right\}.
\end{equation}
Although $y$ can also be obtained via selecting the groups with nonzero coefficients when $\lambda$ in (\ref{E:mspredict}) is chosen properly, we set the threshold $\delta$ as a small positive value to guarantee the robustness to $\lambda$.

Algorithm 2 summarizes the prediction stage of SDGS.
\vspace{-5mm}
\begin{algorithm}[htb]
\SetAlgoLined
\KwIn{$x$, $C^i,i=1,\dots,k$, $\lambda$, $\delta$}
\KwOut{$y$}
Solve group \emph{lasso} in (\ref{E:mspredict}) by using \texttt{group \emph{lasso}}\;
Predict $y$ via thresholding in (\ref{E:thresh})\;
\caption{SDGS Prediction}
\end{algorithm}\vspace{-8mm}


\section{Experiments}

In this section, we evaluate SDGS on several datasets of text classification, image annotation, scene classification, music categorization, genomics and web page classification. We compare SDGS with BR \cite{MLreview2}, ML-KNN \cite{MLKNN} and MDDM \cite{MDDM} on five evaluation metrics for evaluating the effectiveness, as well as the CPU seconds for evaluating the efficiency. All the experiment are run in MatLab on a server with dual quad-core 3.33 GHz Intel Xeon processors and 32 GB RAM.

\subsection{Evaluation metrics}

In the experiments of multi-label prediction, five metrics, which are Hamming loss, precision, recall, F1 score and accuracy, are used to measure the prediction performance.

Given two label matrices $Y1,Y2\in\{0,1\}^{n\times k}$, wherein $Y1$ is the real one an $Y2$ is the prediction one, the Hamming loss measures the recovery error rate:
\begin{equation}\label{E:HammingLoss}
HamL=\frac{1}{nk}\sum\limits_{i=1}^n\sum\limits_{j=1}^k {Y1}_{ij}\oplus{Y2}_{ij},
\end{equation}
where $\oplus$ is the XOR operation, a.k.a. the exclusive disjunction.

The other four metrics are precision, recall, F1 score and accuracy and are defined as:
\begin{align}
&Prec=\frac{1}{n}\sum\limits_{i=1}^n \frac{{\rm card}\left({Y1}_{i}\cap {Y2}_{i}\right)}{{\rm card}\left(Y2_i\right)},\\
&Rec=\frac{1}{n}\sum\limits_{i=1}^n \frac{{\rm card}\left({Y1}_{i}\cap {Y2}_{i}\right)}{{\rm card}\left(Y1_i\right)},\\
&F1=\frac{1}{n}\sum\limits_{i=1}^n \frac{2{\rm card}\left({Y1}_{i}\cap {Y2}_{i}\right)}{{\rm card}\left(Y1_i\right)+{\rm card}\left(Y2_i\right)},\\
&Acc=\frac{1}{n}\sum\limits_{i=1}^n \frac{{\rm card}\left({Y1}_{i}\cap {Y2}_{i}\right)}{{\rm card}\left({Y1}_{i}\cup {Y2}_{i}\right)}.
\end{align}

\subsection{Datasets}

We evaluate the prediction performance and time cost of SGDS on 11 datasets from different domains and of different scales, including Corel5k (image), Mediamill (video), Enron (text), Genbase (genomics), Medical (text), Emotions (music), Slashdot (text) and $4$ sub datasets selected in Yahoo dataset (web data). These datasets were obtained from Mulan's website \footnote{\texttt{http://mulan.sourceforge.net/datasets.html}} and MEKA's website \footnote{\texttt{http://meka.sourceforge.net/}}. They were collected from different practical problems. Table \ref{Table:datasets} shows the number of samples $n$ (training samples+test samples), number of features $p$, number of labels $k$, and the average cardinality of all label vectors $Card$ of different datasets.

\begin{table}[H]
\caption{Information of datasets that are used in experiments of MS. In the table, $n$ (training samples+test samples) is the number of samples, $p$ is the number of features, $k$ is the number of labels, ``Card'' is the average cardinality of all label vectors.}
\begin{center}\vspace{-1mm}
\begin{tabular}{l*{4}{c}}
\hline
Datasets & $n$ & $p$ & $k$ & Card \\
\hline \hline
Corel5k                 & $4500+500$    & $499$     & $374$ & $3.522$  \\
Mediamill               & $30993+12914$   & $120$     & $101$ & $4.376$  \\
Enron                   & $1123+579$    & $1001$    & $53$  & $3.378$  \\
Genbase                 & $463+199$     & $1186$    & $27$  & $1.252$  \\
Medical                 & $333+645$     & $1449$    & $45$  & $1.245$  \\
Emotions                & $391+202$     & $72$      & $6$   & $1.869$  \\
Slashdot                & $2338+1444$    & $1079$    & $22$  & $1.181$ \\
Yahoo-Arts              & $2000+3000$    & $462$     & $26$  & $1.636$ \\
Yahoo-Education         & $2000+3000$    & $550$     & $33$  & $1.461$ \\
Yahoo-Recreation        & $2000+3000$    & $606$     & $22$  & $1.423$ \\
Yahoo-Science           & $2000+3000$    & $743$     & $40$  & $1.451$ \\
\hline
\end{tabular}
\end{center}\vspace{-4mm}
\label{Table:datasets}
\end{table}

\subsection{Performance comparison}

We show the prediction performance and time cost in CPU seconds of BR, ML-KNN, MDDM and SDGS in Table \ref{Table:exp} and Table \ref{Table:yahoo}. In BR, we use the MatLab interface of LIBSVM 3.0 \footnote{\texttt{http://www.csie.ntu.edu.tw/\~cjlin/libsvm/}} to train the classic linear SVM classifiers for each label. The parameter $C\in\left\{10^{-3},10^{-2},0.1,1,10,10^2,10^3\right\}$ with the best performance was used. In ML-KNN, the number of neighbors was $30$ for all the datasets.

\begin{table}[H]
\caption{Prediction performances (\%) and CPU seconds of BR \cite{MLreview2}, ML-KNN \cite{MLKNN}, MDDM \cite{MDDM} and SDGS on Yahoo.}
\begin{center}
\begin{tabular}{|c|l|*{6}{c}|}
\hline
& Methods & Hamming loss & Precision & Recall & F1 score & Accuracy & CPU seconds \\
\hline
\multirow{4}{*}{Arts}&BR & $5$	&$76$	&$25$	&$26$	&$24$	&$46.8$ \\
&ML-knn & $6$	&$62$	&$7$	&$25$	&$6$	&$77.6$ \\
&MDDM & $6$	&$68$	&$6$	&$21$	&$5$	&$37.4$ \\
&SDGS & $9$	&$35$	&$40$	&$31$	&$28$	&$11.7$ \\
\hline
\multirow{4}{*}{Education}&BR & $4$	&$69$	&$27$	&$28$	&$26$	&$50.1$ \\
&ML-knn & $4$	&$58$	&$6$	&$31$	&$5$	&$99.8$ \\
&MDDM & $4$	&$59$	&$5$	&$26$	&$5$	&$45.2$ \\
&SDGS & $4$	&$41$	&$35$	&$32$	&$29$	&$12.6$ \\
\hline
\multirow{4}{*}{Recreation}&BR & $5$	&$84$	&$23$	&$23$	&$22$	&$53.2$ \\
&ML-knn & $6$	&$70$	&$9$	&$23$	&$8$	&$112$ \\
&MDDM & $6$	&$66$	&$7$	&$18$	&$6$	&$41.9$ \\
&SDGS & $7$	&$41$	&$49$	&$36$	&$30$	&$19.1$ \\
\hline
\multirow{4}{*}{Science}&BR & $3$	&$79$	&$19$	&$19$	&$19$	&$84.9$ \\
&ML-knn & $3$	&$59$	&$4$	&$20$	&$4$	&$139$ \\
&MDDM & $3$	&$66$	&$4$	&$19$	&$4$	&$53.0$ \\
&SDGS & $5$	&$31$	&$39$	&$29$	&$26$	&$20.1$ \\
\hline
\end{tabular}
\end{center}
\label{Table:yahoo}
\end{table}

In MDDM, the regularization parameter for uncorrelated subspace dimensionality reduction was selected as $0.12$ and the dimension of the subspace was set as $20\%$ of the dimension of the original data. In SDGS, we selected $r^i$ as an integer in $\left[1,6\right]$, $K\in\left[10^{-6},10^{-3}\right]$, $\lambda\in\left[0.2,0.45\right]$ and $\delta\in\left[10^{-4},10^{-2}\right]$. We roughly selected $4$ groups of parameters in the ranges for each dataset and chose the one with the best performance on the training data. Group \emph{lasso} in SDGS can be solved by many convex optimization methods, e.g., submodular optimization \cite{BachGL} and SLEP \cite{SLEP}. We use SLEP in our experiments.

The experimental results show that SDGS is competitive on both prediction performance and speed, because it explores label correlations and structure without increasing the problem size. In addition, the bilateral random projections further accelerate the computation. SDGS has smaller gaps between precision and recall on different tasks than other methods, and this implies it is robust to the imbalance between positive and negative samples.

\section{Conclusion}

In this paper, we propose a novel multi-label learning method ``Structured Decomposition + Group Sparsity (SDGS)''. Its training stage decomposes the training data as the sum of several low-rank components $L^i_{\Omega_i}$ corresponding to their labels and a sparse residual $S$ that cannot be explained by the given labels. This structured decomposition is accomplished by a bilateral random projections based alternating minimization, and it converges to a local minimum. The row space $C^i$ of $L^i_{\Omega_i}$ is the mapping of label $i$ in the feature subspace. The prediction stage estimates the group sparse representation of a new sample on the multi-subspace $C^i$ via group \emph{lasso}. SDGS predicts the labels by selecting the groups that the nonzero representation coefficients concentrate on.

SDGS finds the mapping of labels in the feature space, where the label correlations are naturally preserved in the corresponding mappings. Thus it explores the label structure without increasing the problem size. SDGS is robust to the imbalance between positive and negative samples, because it uses group sparsity in the multi-subspace to select the labels, which considers both the discriminative and relative information between the mappings of labels in feature subspace.

\bibliographystyle{splncs}
\bibliography{MSRef}

\newpage

\begin{table}[H]
\caption{Prediction performances (\%) and CPU seconds of BR \cite{MLreview2}, ML-KNN \cite{MLKNN}, MDDM \cite{MDDM} and SDGS on 7 datasets.}
\begin{center}\vspace{-2mm}
\begin{tabular}{|c|l|*{6}{c}|}
\hline
& Methods & Hamming loss & Precision & Recall & F1 score & Accuracy & CPU seconds \\
\hline
\multirow{4}{*}{Genbase}&BR & $31$	&$5$	&$39$	&$9$	&$5$	&$1.99$ \\
&ML-knn & $0.6$	&$100$	&$50$	&$92$	&$50$	&$9.38$ \\
&MDDM & $0.6$	&$98$	&$51$	&$92$	&$51$	&$6.09$ \\
&SDGS & $2$	&$83$	&$96$	&$86$	&$70$	&$8.62$ \\
\hline
\multirow{4}{*}{Mediamill} &BR & $4$	&$69$	&$35$	&$43$	&$33$	&$120141$ \\
&ML-knn & $3$	&$41$	&$6$	&$54$	&$5$	&$5713$ \\
&MDDM & $3$	&$36$	&$5$	&$53$	&$4$	&$48237$ \\
&SDGS & $3$	&$58$	&$78$	&$53$	&$37$	&$1155$ \\
\hline
\multirow{4}{*}{Emotions}&BR & $29$	&$55$	&$53$	&$51$	&$42$	&$0.68$ \\
&ML-knn & $28$	&$68$	&$28$	&$41$	&$22$	&$0.66$ \\
&MDDM & $29$	&$54$	&$28$	&$41$	&$22$	&$0.66$ \\
&SDGS & $22$	&$40$	&$100$	&$52$	&$37$	&$0.01$ \\
\hline
\multirow{4}{*}{Enron}&BR & $6$	&$51$	&$28$	&$35$	&$24$	&$77.1$ \\
&ML-knn & $5$	&$51$	&$7$	&$46$	&$5$	&$527$ \\
&MDDM & $5$	&$50$	&$8$	&$49$	&$7$	&$29$ \\
&SDGS & $6$	&$44$	&$50$	&$40$	&$28$	&$271$ \\
\hline
\multirow{4}{*}{Medical}&BR &$31$	&$2$	&$26$	&$5$	&$2$	&$4.88$ \\
&ML-knn &$2$	&$75$	&$7$	&$48$	&$6$	&$22.8$ \\
&MDDM &$2$	&$74$	&$3$	&$30$	&$2$	&$32.3$ \\
&SDGS &$8$	&$36$	&$90$	&$45$	&$26$	&$7.5$ \\
\hline
\multirow{4}{*}{Slashdot}&BR & $13$	&$11$	&$22$	&$14$	&$10$	&$140$ \\
&ML-knn & $4$	&$71$	&$10$	&$31$	&$8$	&$708$ \\
&MDDM & $5$	&$39$	&$1$	&$4$	&$1$	&$114$ \\
&SDGS & $8$	&$38$	&$61$	&$37$	&$27$	&$175$ \\
\hline
\multirow{4}{*}{Corel5k} &BR & $8$ & $2$ & $20$ & $4$ & $2$ & $2240$ \\
&ML-knn & $0.9$	&$62$	&$1$	&$3$	&$0.9$	&$2106$ \\
&MDDM & $0.9$	&$62$	&$1$	&$7$	&$1$	&$458$ \\
&SDGS & $2$	&$9$	&$11$	&$8$	&$5$	&$1054$ \\
\hline
\end{tabular}
\end{center}\vspace{-3mm}
\label{Table:exp}
\end{table}

\end{document}